%% file: main.tex
\documentclass{article}

\usepackage[preprint]{neurips_2026}

\usepackage[utf8]{inputenc} 
\usepackage[T1]{fontenc}    
\usepackage{hyperref}       
\usepackage{url}            
\usepackage{booktabs}       
\usepackage{amsfonts}       
\usepackage{nicefrac}       
\usepackage{microtype}      
\usepackage{xcolor}  
\usepackage{booktabs}
\usepackage{multirow}
\usepackage[table]{xcolor} 
\usepackage{graphicx}
\usepackage{subcaption} 
\usepackage{xspace}
\usepackage{makecell}
\usepackage{threeparttable}
\usepackage{amsmath}
\usepackage{wrapfig2}
\usepackage{siunitx}

\usepackage{algorithm}
\usepackage{algpseudocode}    

\definecolor{customblue}{RGB}{135,200,235}
\newcommand{\ours}[1]{\cellcolor{customblue!30}#1}
\newcommand{\others}[1]{\cellcolor{gray!10}#1}
\newcommand{\methodname}{KVarN\xspace}

\title{\methodname: Variance-Normalized KV-Cache Quantization Mitigates Error Accumulation in Reasoning Tasks}

\author{
Lorenz K. Muller\thanks{Correspondence to: Lorenz K. Muller <lorenz.mueller@huawei.com>.} \\
Huawei
\And Philippe Bich \\
Huawei
\And Chiara Boretti \\
Huawei
\AND Hyun-Min Chang \\
Huawei
\And Jiawei Zhuang \\
Huawei
\And Lukas Cavigelli \\
Huawei
}

\begin{document}

\maketitle

\begin{abstract}
Test-time scaling is a powerful approach to obtain better reasoning in large language models, but it becomes memory-bottlenecked during long-horizon decoding, as the KV-cache grows. KV-cache quantization can help improve this, but current methods are evaluated under prefill-like settings and errors behave differently under autoregressive decoding. We show that in the latter regime, quantization errors accumulate across timesteps, driven primarily by incorrect token scales. We introduce KVarN, a calibration-free KV-cache quantizer that applies a Hadamard rotation followed by a dual-scaling variance normalization across both axes of the $K$ and $V$ matrices. We find that this combination fixes outlying token-scale errors and substantially reduces error accumulation over existing baselines. KVarN establishes a new state-of-the-art for KV-cache quantization on generative benchmarks, including MATH500, AIME24 and HumanEval, at 2-bit precision. A vLLM implementation of the KVarN method is available at \url{https://github.com/huawei-csl/KVarN}.
\end{abstract}

\section{Introduction}
Recently,  test-time scaling \cite{openai2024learning} has helped reasoning LLMs achieve new levels of competence in a variety of domains \cite{snell2025scaling, pmlr-v267-bi25a-Forest-of-Thought, muennighoff2025s1, shen2026thinking}. As the name implies, the premise of test-time scaling is an ever-increasing decoding length that supports this greater capability. Consequently, the efficient handling of the KV-Cache is becoming more and more relevant to achieve optimal reasoning per time trade-offs. 

One avenue of attack on this memory bottleneck is KV-Cache quantization. Such methods try to assign fewer than 16 bits per element of the $K$ and $V$ matrices, often as few as 2 to 4. With effective reshaping and blocking strategies, linear methods like KIVI \cite{KIVI} can already achieve good results, and codebook-based methods like TurboQuant \cite{zandieh2026turboquant} can further improve on them. 
Typical evaluation settings for these methods are large prefill problems, where a fixed, long context is quantized in parallel. Examples are Needle-in-a-Haystack (NiaH) \cite{kamradt2023needle}, MultiQA \cite{talmor-berant-2019-multiqa}, or LongBench \cite{bai2024longbench}. 

However, in long-form test-time scaling scenarios, the generated token sequence needs to be compressed on-the-fly as it is produced. As we will show in this paper, this leads to the accumulation of quantization error across decoding timesteps, because standard methods fail to preserve per-token scales during quantization. 

One highly successful approach in weight quantization in recent years has been incoherence processing \cite{chee2023quip}, e.g., by Hadamard rotation. We elucidate that this approach alone fails for KV-Cache compression, because it does not sufficiently address token scaling. We identify dual-scaling with Sinkhorn-based variance-normalization as an effective approach to further mitigate token scaling errors. 

\subsection{Contributions}

\begin{itemize}
    \item End-to-end evaluations of \underline{K}V-Cache quantization with \underline{Var}iance \underline{N}ormalization (\methodname \footnote{kvarn, noun, Swedish: the grinding apparatus for substances such as grains, seeds, spices, coffee beans, KV-Caches, etc.}) on generative benchmarks with \textbf{substantial improvement over current state-of-the-art} in AIME24, MATH500, HumanEval and IFEval.
    \item Identifying token magnitude errors as a key driver of outlier errors.
    \item Showing that outlier errors contribute disproportionally to end-to-end quality.
    \item An efficient pseudo-decode evaluation method for error accumulation in KV-Cache quantization relevant for test-time scaling.
    \item A novel KV-Cache compression method that mitigates token magnitude errors and error accumulation across timesteps called \methodname.
    
\end{itemize}

\begin{figure}
    \centering

    \begin{subfigure}[b]{0.414\linewidth}
        \centering
        \includegraphics[width=\linewidth]{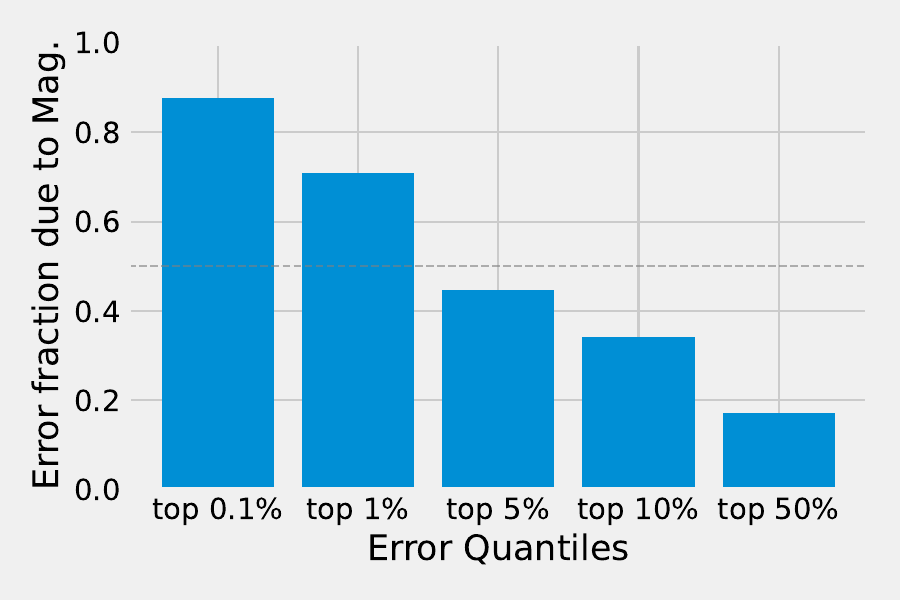}
        \caption{}
        \label{fig:scale:a}
    \end{subfigure}
    \hfill
    \begin{subfigure}[b]{0.57\linewidth}
        \centering
        \includegraphics[width=\linewidth]{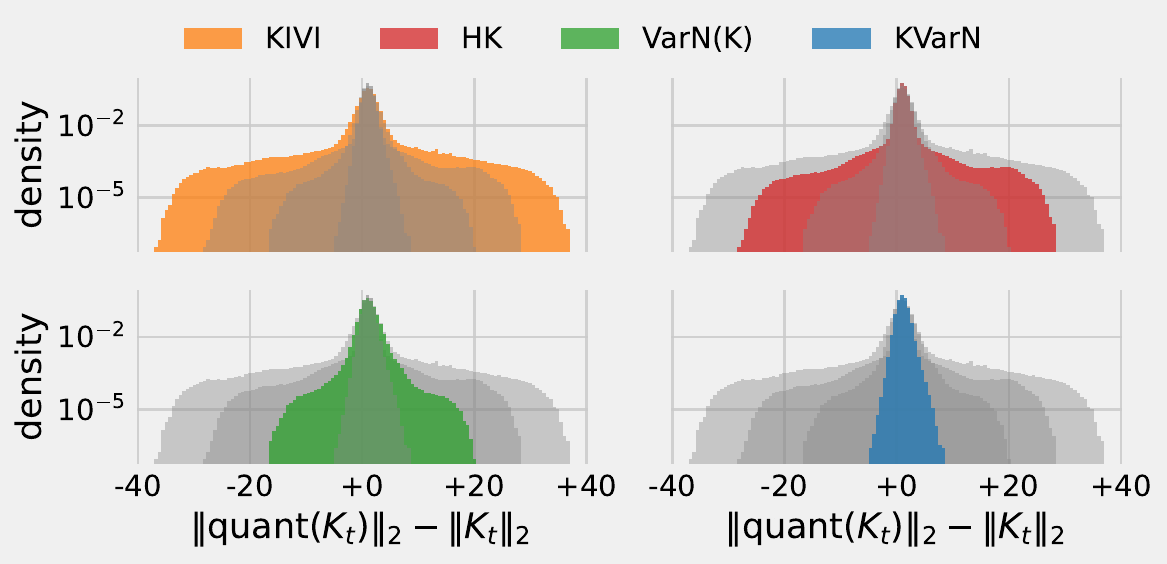}
        
        \caption{}
        \label{fig:scale:b}
    \end{subfigure}
    \hfill
    
    \caption{(\subref{fig:scale:a}) What fraction of the top k\% largest errors is due to magnitude rather than direction (decomposed as in Eq.~\ref{eq:decompose}). Large quantization errors in $K$ are mostly due to incorrect token scaling. To fix outlier errors, the token magnitude needs to be better preserved.  (\subref{fig:scale:b}) Difference of per-token magnitude of quantized $K$ to full-precision $K$ matrix on Qwen3-4B under different 2-bit quantization methods (KIVI \cite{KIVI}, HK: Hadamard rotated $K$, VarN(K): Variance-Normalized $K$, and \methodname: our proposed method). Variance normalization prevents the rounding process from scaling the norm of worst-case tokens; the effect synergizes with scale-error reduction by Hadamard rotation. \methodname effectively suppresses magnitude errors, which leads to better end-to-end test-time scaling.}
    \label{fig:token-norm}
\end{figure}






\section{Preliminaries}

Unless otherwise mentioned, illustrative figures use the KV-Cache of Qwen3-4B~\cite{yang2025qwen3technicalreport} ingesting wikitext-2 subsets as underlying data. Generally $K$ quantization is often considered to be more difficult than $V$ quantization\footnote{As an illustration, in Llama3.1-8B $>98\%$ of the top 5\% quantization errors under the KIVI-scheme lie in a $K$-matrix.}; we will also focus more discussion on $K$ because of this.  

\subsection{Basics of KV-Cache Quantization}
Following previous KV-Cache quantization work \cite{KIVI}, we process the KV-cache in the channel dimension per-head and in the token dimension in chunks (e.g. of size 128). Consequently, the base-tile to work with is a tile of shape ($\text{head-dim}\times \text{token-chunk}$), e.g. for Llama3.1-8B in our setup a $128\times128$ tile. KIVI~\cite{KIVI} has shown that with round-to-nearest quantization, it is best to  quantize the $V$ matrix per token and the $K$ matrix per channel. E.g., the $K$  matrix is quantized to a set consisting of a low-precision matrix $K_q$, and two high-precision vectors, containing one element for each channel. One of these vectors is an offset (or zeropoint) $\vec{z}$ the other a scale $\vec{s}$. The dequantized approximation of $K$ is $K_{dq}$ obtained by
\begin{equation}
    K_{dq} = (K_q + \vec{z}) \odot \vec{s}. 
\end{equation}

\subsection{Incoherence Processing for Quantization}
\label{sec:rotate}
The distribution of to-be-quantized tensors can often be made more favorable by applying some simple transformations to them \cite{ashkboos2024quarot, liu2025spinquant}. Particularly useful are transformations that can be either applied cheaply on-the-fly or precomputed and absorbed into existing model weights. The Hadamard transform is fast enough for online application (with $O(N \log N)$ complexity) and can often be absorbed into adjacent weight matrices (see Fig.~\ref{fig:figure_att} in the Appendix). In the limit of large matrices, the Hadamard transform yields Gaussian-distributed outputs \cite{malinovskii2025higgs, chee2023quip, tseng2024qtip}. This is often referred to as incoherence processing. We follow the same layout as QuaRot \cite{ashkboos2024quarot}, see Fig.~\ref{fig:figure_att} in the Appendix.

We find that in KV-Cache quantization, incoherence processing is helpful to equalize channel-space outliers, but insufficient on its own to manage token-wise scaling errors, see Fig.~\ref{fig:token-norm}.

\subsection{Dual-Scaling in LLM Weight Quantization}
An alternative preconditioning transform has recently become popular in LLM weight quantization: Scaling both input- and output-channel dimensions to have approximately uniform variance by way of a variance-targeted Sinkhorn-Knopp-style normalization. This normalization is helpful because it approximates calibration data from weight structure. Thereby, it improves end-to-end output quality while it paradoxically causes weight-matrix reconstruction errors to increase \cite{muller2025sinq}.

We find that dual-scaling variance normalization is also helpful in KV-Cache quantization, but for unrelated reasons as there is no calibration data to approximate (see Sec. \ref{sec:sinq-kv}).

\begin{figure}
    \centering
    \includegraphics[width=\linewidth]{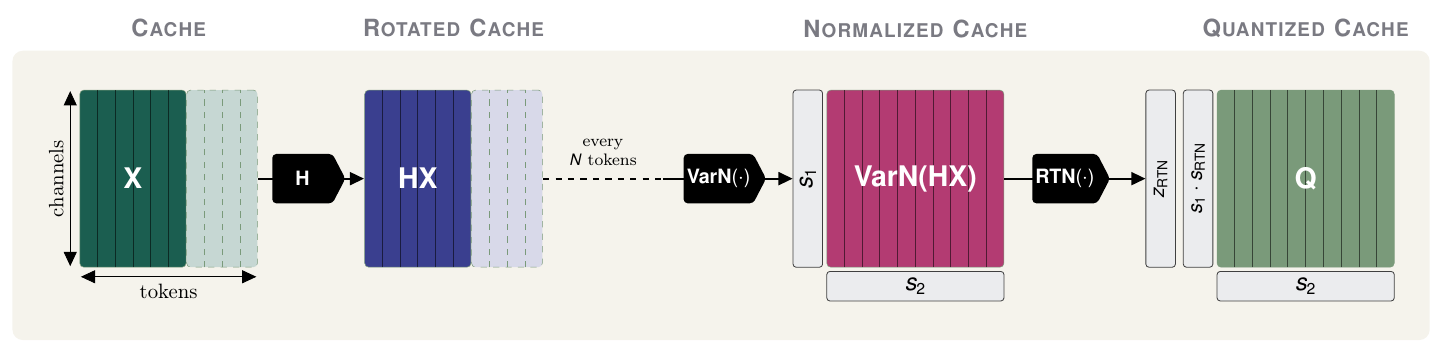}
    \caption{Schematic layout of \methodname. Every token is Hadamard-rotated in the channel dimension. After one block of tokens (e.g., 128) of generation, each block is variance-normalized in token and channel dimension, denoted by VarN($\cdot$). Finally, it is quantized with round-to-nearest (RTN). We store the standard scale and zero-point of the RTN, plus a second scale. At 2.3 average bits per element even with the second scale, \methodname outperforms or matches prior methods, see e.g. Tab.~\ref{tab:ifeval_results_horizontal}.}
    \label{fig:sinkv-pipeline}
\end{figure}

\subsection{Key Ideas}
\begin{wrapfigure}{r}{0.41\textwidth} 
    \vspace{-42pt}
  \begin{center}
    \includegraphics[width=0.36\textwidth]{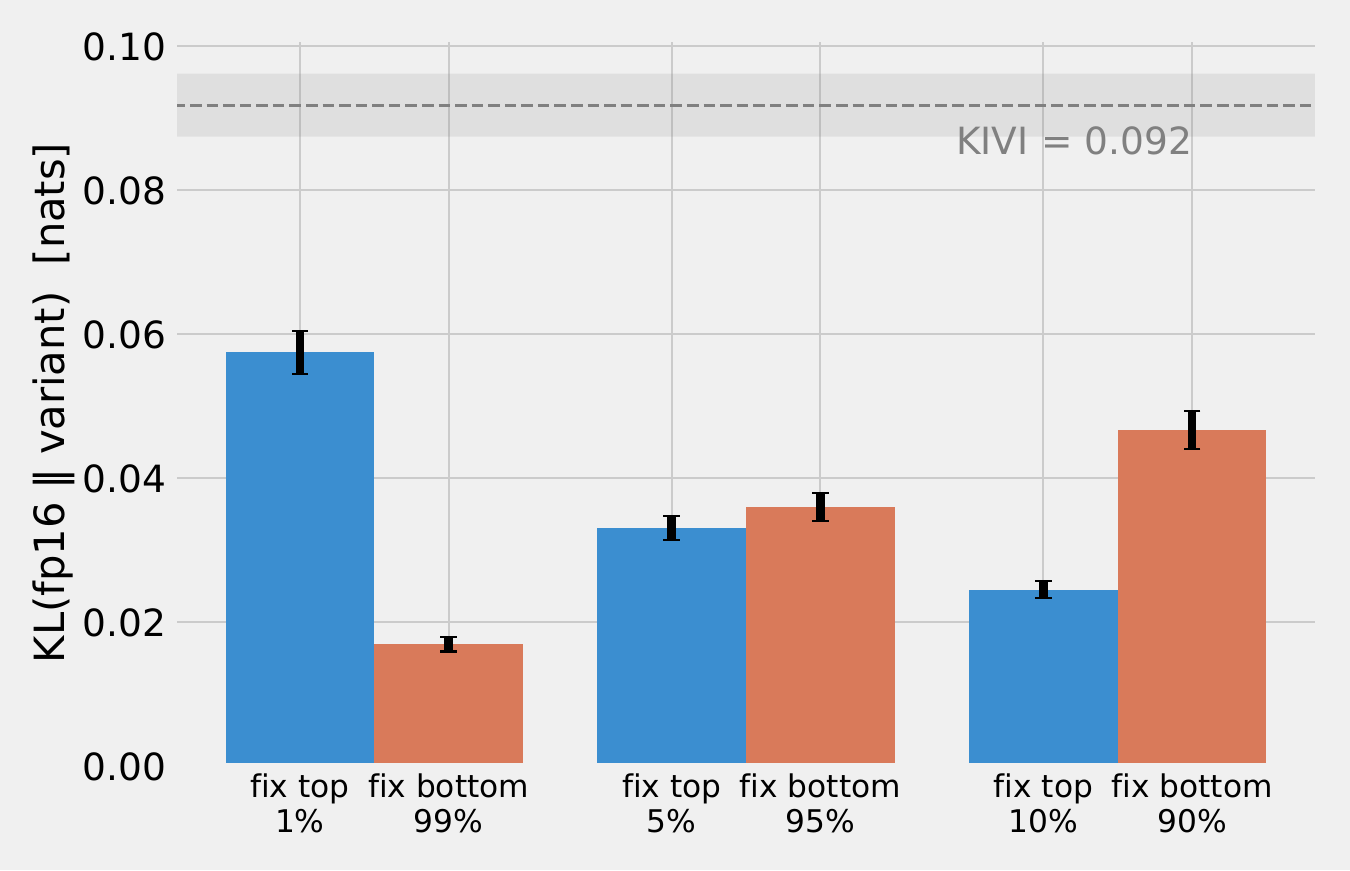}
  \end{center}
  \caption{Replacing the 5\% worst outlier errors with high precision values improves end-to-end KL-divergence more than fixing the other 95\%, even though more MSE lies there (see Fig.~\ref{fig:outlier_MSE}).}
  \label{fig:outlier_impact}
  \vspace{-10pt} 
\end{wrapfigure}

The central argument of this paper is:  The largest, e.g. top 5\%, errors in the KV-Cache cause most of the end-to-end degradation, smaller errors (even if they are many) are less important (see Fig.~\ref{fig:outlier_impact}). The magnitude per token is the main driver of such outlying errors (see Fig.~\ref{fig:scale:a}). \methodname fixes these magnitude errors with combined incoherence processing and dual-scaling (see Fig.~\ref{fig:scale:b}). This also mitigates error accumulation over time (see Figs.~\ref{fig:pseudo-decode} and ~\ref{fig:output_mae_vs_rtn}). In this way, \methodname achieves state-of-the-art end-to-end KV-Cache quantization in reasoning and instruction following. See Tabs.~\ref{tab:math_reasoning_results}, \ref{tab:humaneval_results} and \ref{tab:ifeval_results_horizontal}.

\section{Methods}

\subsection{Distinguishing Magnitude-based and Directional Error}
To isolate the distinct effects of magnitude shift and directional distortion introduced by quantization, we can decompose the squared $\mathcal{L}_2$ error between the full-precision key vector, $K$, and its dequantized counterpart, $K_{dq}$. Using the geometric definition of the dot product, where $\theta$ represents the angle between $K$ and $K_{dq}$, we expand the squared error norm:
\begin{equation}
    \|K - K_{dq}\|^2 = \|K\|^2 - 2\|K\|\|K_{dq}\|\cos\theta + \|K_{dq}\|^2
\end{equation}
By adding and subtracting the cross-term $2\|K\|\|K_{dq}\|$, we can decouple the expression into two distinct components:
\begin{align}
    \underbrace{\|K - K_{dq}\|^2}_{E_T, \text{Total Error}} &= \left( \|K\|^2 - 2\|K\|\|K_{dq}\| + \|K_{dq}\|^2 \right) + \left( 2\|K\|\|K_{dq}\| - 2\|K\|\|K_{dq}\|\cos\theta \right) \nonumber \\
    &= \underbrace{\left( \|K\| - \|K_{dq}\| \right)^2}_{E_M, \text{Magnitude Error}} + \underbrace{2\|K\|\|K_{dq}\|(1 - \cos\theta)}_{E_D, \text{Directional Error}}
    \label{eq:decompose}
\end{align}
In this way we can decompose the total quantization error into a pure magnitude penalty and a pure directional penalty (scaled by the norms and governed by the cosine similarity). This decomposition allows us to independently analyze how $K$-magnitude inflation versus angular quantization noise impacts the attention logits.

In Fig.~\ref{fig:scale:a} we use $\frac{E_M}{E_T}$, the fraction of the total error due to magnitude, to show that outlier errors are overwhelmingly caused by incorrect magnitudes.

\subsection{Evaluation of KV-Cache Quantization Error Accumulation}
\begin{figure}
    \centering
    \includegraphics[width=.9\linewidth]{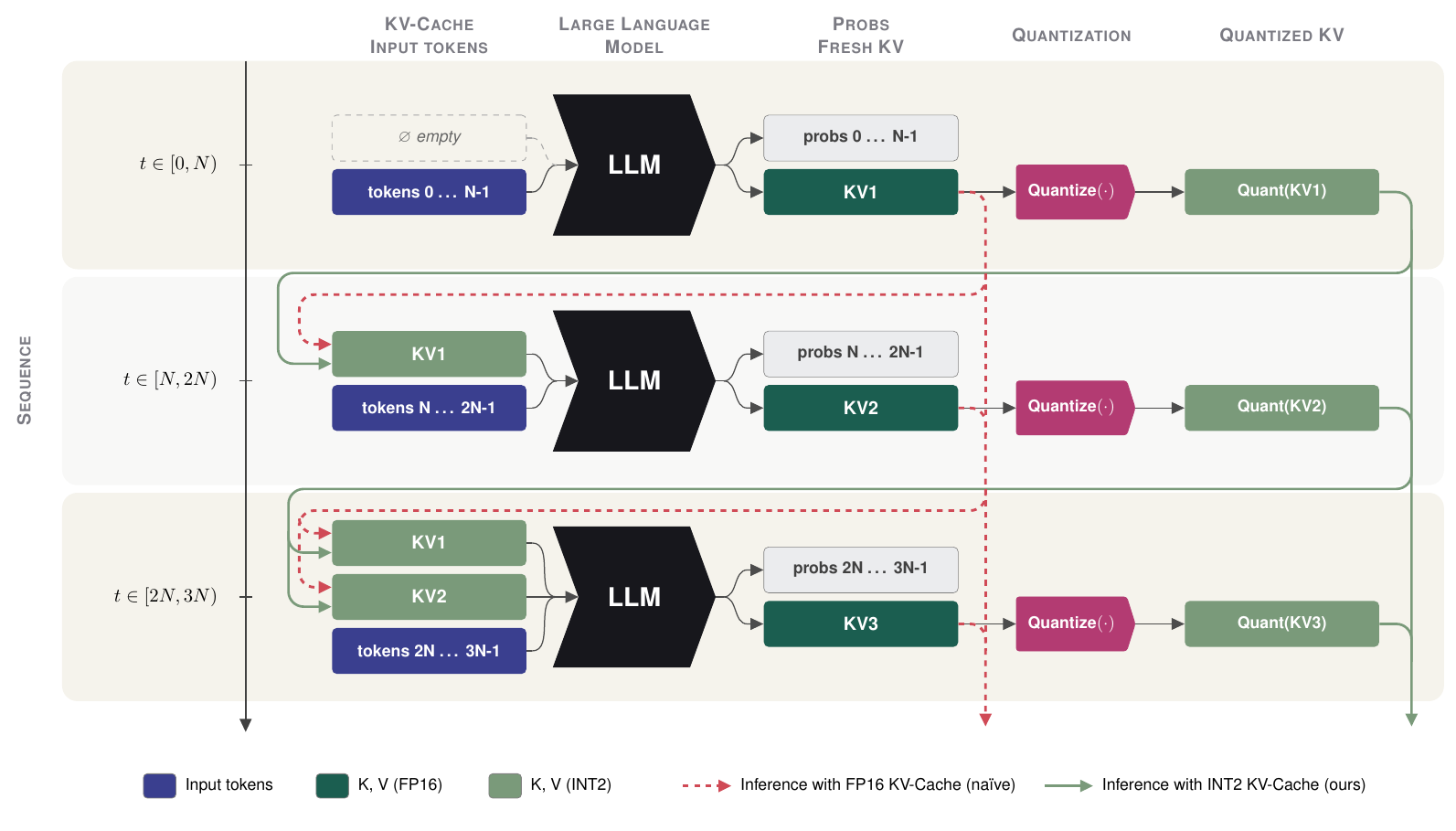}
    \caption{Prior KV-Cache quantization papers address the parallel prefill scenario (red dashed arrows). We propose a `pseudo-decode' setting (green solid arrows) to better model decoding errors. We split the sequence into blocks of size b. After every block, the freshly produced $K$, $V$ are quantized before being written back to the KV-cache. Subsequent blocks access a \emph{quantized} cache, so quantization error accumulates over time. LLMs operate this way during decoding of long sequences (e.g., in test-time scaling, such as reasoning tasks). \methodname is designed to operate in this regime. }
    \label{fig:pseudo-decode}
\end{figure}
\label{sec:err_acc}

When the KV-Cache is quantized during decoding in deep models, a form of error accumulation occurs, see Fig.~\ref{fig:pseudo-decode} for an illustration. Namely, when the attention in transformer block $B_l$ at block-index $l$ is computed with a \emph{quantized} KV-Cache, this error influences the $K$ and $V$ matrices that the attention layer in the subsequent transformer block $B_{l+1}$ produces. So the unquantized $K$ and $V$ matrices at $B_{l+1}$ are already slightly wrong. Quantizing them introduces additional error, which in turn affects $K$ and $V$ at $B_{l+2}$ and beyond. In this way, KV-Cache quantization errors can be accumulated across layers and eventually time-steps. This problem compounds as the generated sequence gets longer.

To directly measure the extent of this problem, we propose a fast decode-like LLM evaluation.  Namely, we divide our full prefill sequence into blocks of size $b$. Whenever $b$ tokens have passed through the model we are evaluating, we quantize the KV-Cache. Then, for all later tokens, we compute latent states \emph{with this quantized KV-Cache} (as would be done at decoding time). We call this the `pseudo-decode' setting. In Fig.~\ref{fig:output_mae_vs_rtn} we show this error accumulation is specifically suppressed by \methodname, which we will introduce below.

\subsection{\methodname: Variance-Normalized Quantization for KV-Cache}
\label{sec:sinq-kv}
In KVarN, we apply two transformations to mitigate token scaling errors: First, rotate in the channel dimension with the Hadamard transform, and second, normalize with variance scaling (online) in both channel and token dimension. See Fig.~\ref{fig:sinkv-pipeline} for an overview. For the Hadamard transformation we follow the now widely-used setup of QuaRot \cite{ashkboos2024quarot}, see Fig.~\ref{fig:figure_att}. This rotates the $K$ and $V$ matrices in the channel dimension and reduces outliers there. 
In the token dimension it would be too costly to apply a Hadamard rotation (because on decompression this rotation will have to be undone online for every token position; i.e. typically $128^2$ operations per 128 tokens per channel).  



Following recent \emph{weight} quantization papers \cite{muller2025sinq}, we apply scaling factors along both dimensions of the to-be-quantized tensor; for $K$ and $V$ these are the channel and token dimensions. Prior KV-Cache quantization methods only scale one dimension. In contrast to Hadamard rotation, the additional element-wise scaling increases the dequantization overhead only by one FLOP per token per channel. The additional scaling allows us to perform the variance normalization below, which we denote by VarN() in the rest of the paper. 

To obtain the two scaling vectors, we iteratively normalize the column-wise and row-wise variance.
It is not possible to work directly with the magnitude of the matrices, because they are signed and in some cases have large offsets. One cannot directly use the token-axis variance alone to normalize tokens, because this would increase the per-channel kurtosis. Iterative variance normalization can avoid this, while still normalizing per token scales. 
This yields $K$ and $V$ matrices with normalized rows and columns. Before quantization, the variance of the token and channel dimensions of these normalized matrices are uniform. In practice, we adapt the log-domain standard-deviation-scaling implementation of SINQ \cite{muller2025sinq} to variance normalization for KV-Caches, see Appendix Sec.~\ref{sec:sinq-alg}, Alg.~\ref{alg:sinq}.

 In weight quantization, normalizing the standard deviation is helpful because it approximates calibration data from weight structure (and paradoxically even causes weight-matrix reconstruction errors to increase) \cite{muller2025sinq}. In our setting, this normalization is useful because it reduces directly the matrix reconstruction error (not because of a fortuitous alignment of rounding errors and typical input data). Specifically, it decreases tail-errors that are primarily due to incorrect token scaling by directly fixing the magnitude with an additional high-precision scale, see Fig.~\ref{fig:scale:b}.


\subsection{Local Proxy Test: Attention Output Reconstruction}
\begin{figure}
    \centering
    \begin{subfigure}[b]{0.3\linewidth}
        \centering
        \includegraphics[width=\linewidth]{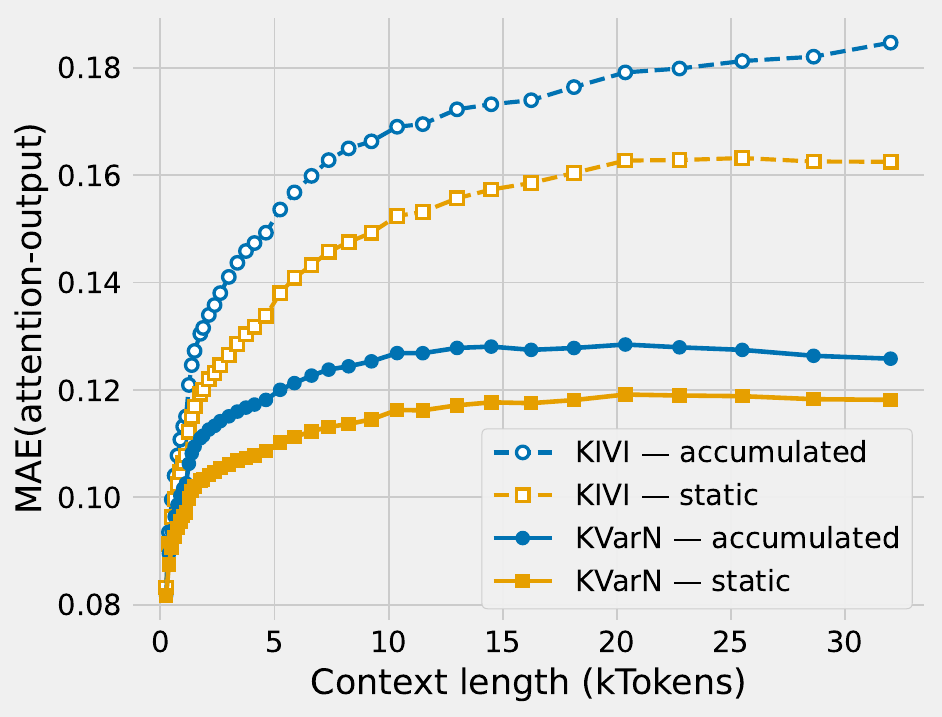}
        \caption{Average attention output reconstruction error}
        \label{fig:sub_a}
    \end{subfigure}
    \hfill
    \begin{subfigure}[b]{0.3\linewidth}
        \centering
        \includegraphics[width=\linewidth]{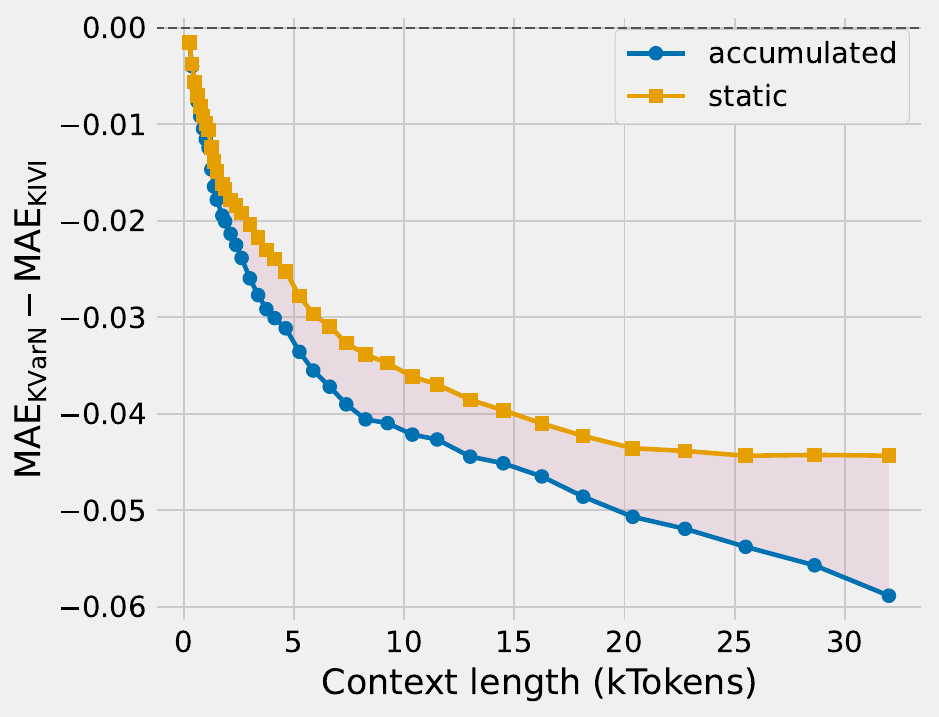}
        \caption{Difference of our method to KIVI}
        \label{fig:sub_b}
    \end{subfigure}
    \hfill
    \begin{subfigure}[b]{0.3\linewidth}
        \centering
        \includegraphics[width=\linewidth]{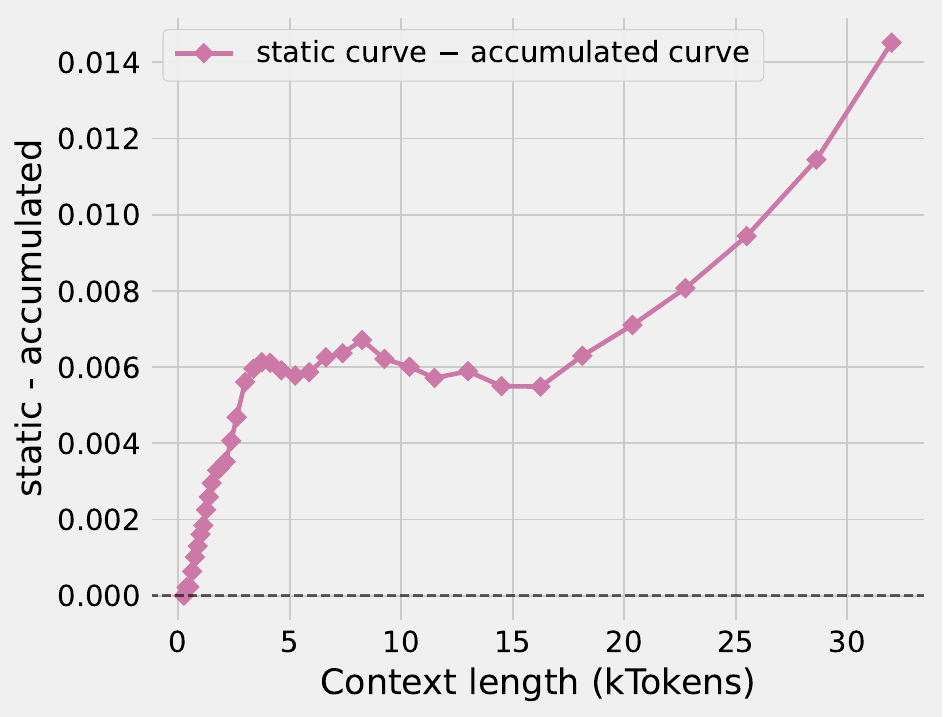} 
        \caption{Difference between the two curves in (\subref{fig:sub_b})}
        \label{fig:sub_c}
    \end{subfigure}
    \caption{Average reconstruction error after quantization across all attention layer outputs in \hbox{Qwen3-4B}. (\subref{fig:sub_a}) Our method has smaller error than KIVI at all context lengths.  As discussed in Sec.~\ref{sec:err_acc} errors accumulate over time. Prior work considers the static case (orange) here we consider the dynamic accumulation (blue) (\subref{fig:sub_b}), (\subref{fig:sub_c}) \methodname has an increasing advantage over KIVI as contexts get longer.}
    \label{fig:output_mae_vs_rtn}
\end{figure}
To illustrate the impact of our method, we consider the norm of the reconstruction error for the output of the attention layer under quantized $K$ and $V$. We evaluate this both under a prefill-like condition that does not consider error-accumulation over quantization steps, and the pseudo-decode setting as given in Sec.~\ref {sec:err_acc}.

In Fig.~\ref{fig:output_mae_vs_rtn} we see that compared to RTN (KIVI), the proposed \methodname achieves much lower error, and accumulates less error over time. \methodname has much lower token scale errors, see Fig.~\ref{fig:scale:b}; in the worst case, such scale errors can compound exponentially (repeated application of incorrect multipliers) and avoiding them is particularly helpful over longer contexts in accumulating regimes. 




\section{Experiments}
\label{sec:exp}

\subsection{Quantization Quality}
\label{sec:quality}
In this section, we consider various metrics for the quality of a KV-cache quantization. We focus here on metrics that reveal error-accumulation effects, see Section \ref{sec:err_acc}. 

We evaluate on Qwen3-4B \cite{yang2025qwen3technicalreport}, Llama-3.1-8B \cite{grattafiori2024llama} and Phi-4-14B \cite{abdin2024phi} to cover a range of model sizes and families. The Qwen model natively supports reasoning, and for the Phi model, there is a reasoning variant. Llama-3.1-8B does not have a reasoning variant. We evaluate all models on line-retrieval and instruction following and the reasoning models on the reasoning benchmarks. Additional experimental details are discussed in Appendix Sec.~\ref{sec:exp_details}.


\subsubsection{End-to-End Reasoning and Instruction-Following Performance}
The target of \methodname are long-context generation tasks like reasoning, coding and instruction following. Specifically, we consider MATH500 \cite{lightman2024lets_math500}, AIME24 \cite{aime24}, HumanEval \cite{chen2021evaluating_humaneval} and IF-Eval \cite{zhou2023instructionfollowingevaluationlargelanguage_ifeval}. MATH500 is a mathematical reasoning benchmark requiring models to formulate complex derivations and output mathematical solutions.  AIME24 similarly tests competition-level mathematical capabilities. Answers typically require long-horizon chains of thought, and exact integer answers. HumanEval assesses programming proficiency by measuring a model's ability to translate natural language docstrings into functionally correct Python code. Finally, the Instruction Following Evaluation benchmark IF-Eval tests a model's ability to  adhere to formatting and content constraints. It measures reliability in executing structural directives, e.g., word-count limits or specific output templates. We report average accuracy on these benchmarks over 3 runs. \methodname achieves the best overall performance with the lowest average bits (2.3 per element of KV-Cache), see Tabs.~\ref{tab:math_reasoning_results}, \ref{tab:humaneval_results}, \ref{tab:ifeval_results_horizontal}.

\input{tables/reasoning_appendix}
\input{tables/humaneval_appendix}
\input{tables/ifeval_main}




\subsubsection{Line-Retrieval Accuracy}
Because synthetic tasks, especially NiaH \cite{kamradt2023needle}, are commonly used in prior work, we add line-retrieval \cite{li2023how_line-retrieval}.
While similar to NiaH we find line-retrieval to be more informative. NiaH seems comparatively too easy, especially, when error accumulation across time is not taken into account.  See Appendix Sec.~\ref{sec:niah} for NiaH evaluations. We give comprehensive results on line-retrieval for various baselines and models, see Tab.~\ref{tab:kv_quant_results}. \methodname performs best overall. 

\input{tables/line_retrieval_main}


\subsection{Runtime Overhead}
In \methodname whenever a new chunk of KV-Cache is stored (e.g., every 128 tokens), it needs to be normalized with iterative variance-scaling. This raises the question: Does this online rescaling operation cause a significant overhead?

\begin{figure}[ht]
    \centering
    \includegraphics[width=0.9\linewidth]{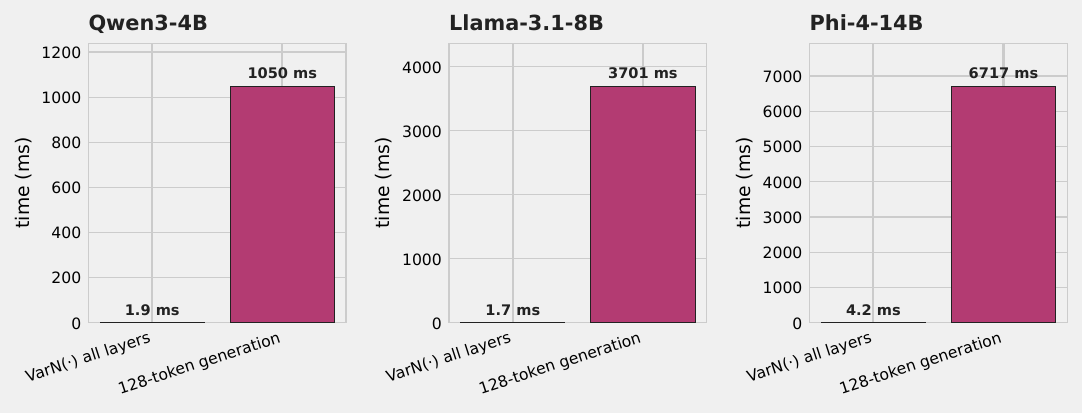}
    \caption{Speed measurement on GPU in the fast vLLM framework. The variance-normalization causes a very minor overhead.}
    \label{fig:timing}
\end{figure}

To answer this question, we compare the time taken to generate 128 tokens to the time required for 8 iterations of variance-normalization in every attention layer in Qwen3-4B. The base-model is run through vLLM \cite{kwon2023efficientvLLM} at fp16. On the same hardware (a GPU with 500 TFLOP at fp16 and 1.8 TB/s memory bandwidth) the optimized vLLM token generation takes 1050 ms, while the proposed normalization takes 1.9 ms, see Fig.~\ref{fig:timing}. This is a 0.18\% measured overhead compared to standard methods like KIVI \cite{KIVI} for the quantization. For larger models the relative overhead decreases. 

The dequantization operation with two scales instead of one as used in KVarN is known to cause a minor slow-down of about 1\% compared to RTN~\cite{muller2025sinq} (see Appendix Sec.~\ref{sec:kvarn_dequant}). Most prior methods have larger overhead during dequantization, e.g. for code-book look-ups or for handling mixed-precision. 




\section{Related Work}

\subsection{KV-Cache Quantization}
Most closely related to our approach are works on KV cache quantization, which aim to reduce memory usage by converting full-precision keys and values to low-bit representations while preserving model performance. Early methods such as KIVI \cite{KIVI} observe that keys and values exhibit distinct statistical properties, and therefore propose treating them differently, quantizing keys per channel and values per token to better capture their variability. Building on this idea, KVQuant \cite{hooper2024kvquant} introduces non-uniform quantization combined with outlier-aware handling, retaining a small fraction of high-magnitude channels (around 1\%) in full precision to maintain accuracy.
Other approaches like Kitty \cite{xia2025kitty} proposes a 2-bit quantization scheme augmented with channel-wise importance selection, where a subset of key channels (e.g., the top ~12\%) is stored at higher precision (4-bit). In a different direction, PolarQuant \cite{wu2026polarquant} represents keys in polar coordinates, quantizing radius and angle separately while applying more aggressive compression to values.

Recently, TurboQuant \cite{zandieh2026turboquant} frames KV cache compression as a near-optimal vector quantization problem. It first applies a random rotation to the KV representations, redistributing information more uniformly across dimensions and making simple scalar quantization nearly optimal. Then it further refines the process through a lightweight residual correction based on a 1-bit quantized Johnson–Lindenstrauss transform to better preserve inner products. 

Our approach builds primarily on KIVI \cite{KIVI}, where we identify the token scaling problem and address it with \methodname. We  find that MSE-optimal quantization is not optimal end-to-end, because outliers are of disproportionate importance.

\subsection{KV-Cache Compression with Eviction and/or Token-Merging}
KV cache compression can be also achieved by deleting or merging tokens. Eviction operates by pruning KV cache entries from the prefill stage at token level. H2O \cite{NEURIPS2023_6ceefa7b_H2O} and Scissorhands \cite{NEURIPS2023_a452a7c6_scissorhands} use attention scores to estimate token importance, keeping entries that receive high attention from subsequent queries. SnapKV \cite{NEURIPS2024_28ab4182_SnapKV} and PyramidKV \cite{cai2025pyramidkv} refine the selection strategy by computing attention-based importance scores over KV pairs using queries from a trailing context window, incorporating future information into the eviction decision. 
In contrast, KVZip \cite{kim2026kvzip} adopts a query-agnostic perspective by evaluating the importance of each entry based on its contribution to a reconstruction of the original context through a teacher-forced decoding task.

Complementing the eviction-based approaches, token merging reduces the number of tokens by combining them into a smaller set of informative representation. Recent works, such as CaM: Cache Merging for Memory-efficient LLMs Inference \cite{zhang2024cam} applies a Bernoulli-based masking process to merge value states within the KV cache during long-sequence generation. The merging strategy applies uniformly across all layers, ignoring varying attention patterns across layers.
D2O \cite{wan2025textdtexto_d2o} proposes a KV-cache-level merging strategy to overcome this. It dynamically allocates cache capacity across layers and adapts the merging process using an exponential moving average (EMA) threshold, reducing information loss and improving efficiency in autoregressive generation tasks.

Token-merging and eviction are largely orthogonal to quantization and can be combined with it. However, for completeness we add comparisons to such methods in the Appendix, see Sec.~\ref{sec:evict}.


\section{Conclusion}
In this paper we have shown that under current KV-Cache quantization methods, end-to-end output degradation is driven by outlier errors that are brought about by incorrect token scaling. 
We address this scaling with \methodname, which excels at suppressing error accumulation that occurs in long decoding tasks under KV-Cache quantization.
With \methodname we achieve state-of-the-art quality on generative benchmarks including AIME24, MATH-500, HumanEval and IF-Eval, enabling near loss-less 2.3bit-per-element KV-Caches with a measured quantization latency overhead of $0.18\%$ over baseline.

\bibliographystyle{plain}

\input{main.bbl}
\newpage
\appendix

\section{Hadamard Transform Arrangement}
\begin{figure}
    \centering
    \includegraphics[width=\linewidth]{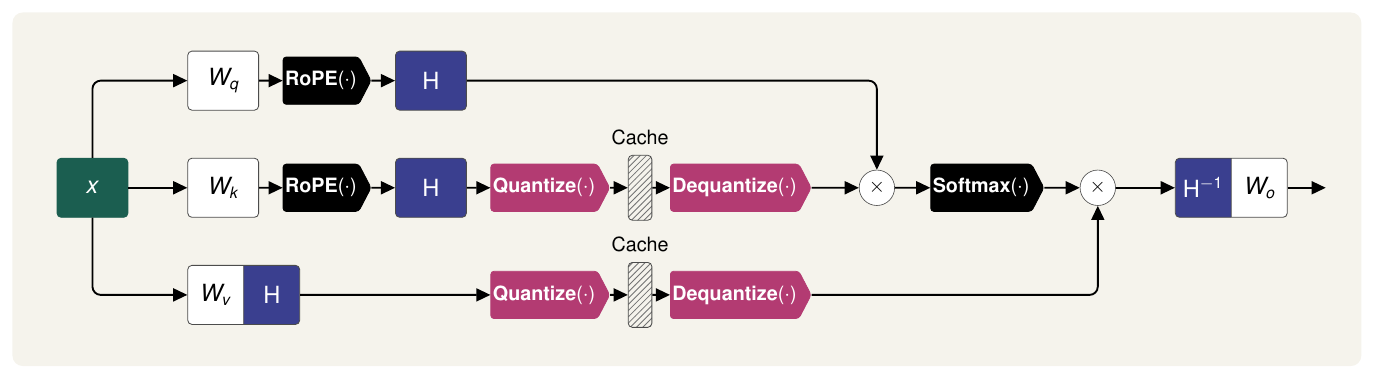}
    \caption{Arrangement of Hadamard transforms in an attention layer in our method.}
    \label{fig:figure_att}
\end{figure}

The arragement of Hadamard transforms we use is similar to QuaRot \cite{ashkboos2024quarot}, see Fig.~\ref{fig:figure_att}. There are two absorbed transforms (that are merged with $W_V$ and $W_O$ respectively and do not need to be computed during inference) and two online transforms after the RoPE-embedding. The two online transforms are head-wise, i.e. seen on the full layer they are block-diagonal Hadamard transforms with a block-size equal to the head-size. The complexity of these online transforms is $O(N \log N)$, which is negligible compared to the linear layers next to them \cite{ashkboos2024quarot}.

\section{Full Token-Scale Distribution Error}

\begin{figure}
    \centering
    \includegraphics[width=\linewidth]{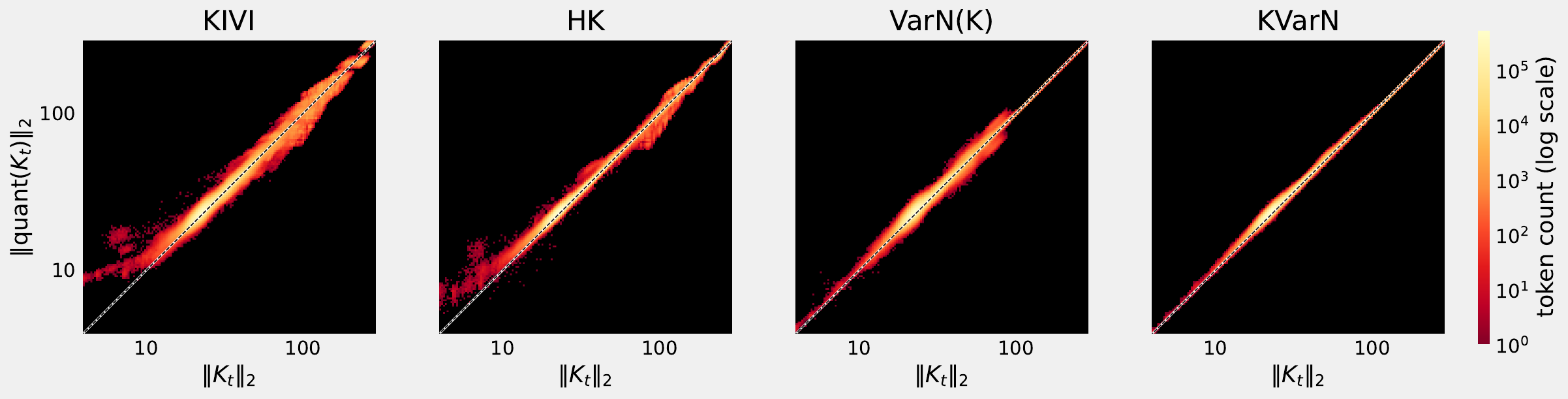}
    \caption{Joint distribution of $K$ magnitude and quantized $K$ magnitude using different quantization methods. \methodname tightly controls token scales, while the baseline KIVI and ablations of our method have substantial off-diagonals.}
    \label{fig:joint}
\end{figure}

We provide the full empirical joint distribution of per token $K$ magnitude before and after quantization in Fig.~\ref{fig:joint} using different ablations of our method. KVarN successfully contains the magnitudes to the diagonal. We see that Hadamard rotation and variance-normalization have different, complementary effects: Hadamard rotation squeezes the distribution close to the identity, but at very large and very small tokens performs much worse than variance normalization. Variance Normalization in contrast is especially effective at these extremes and in KVarN we clearly see the synergistic effect of both in \methodname.

\section{Outlier Contributions to MSE}
\begin{figure}
    \centering
    \includegraphics[width=0.5\linewidth]{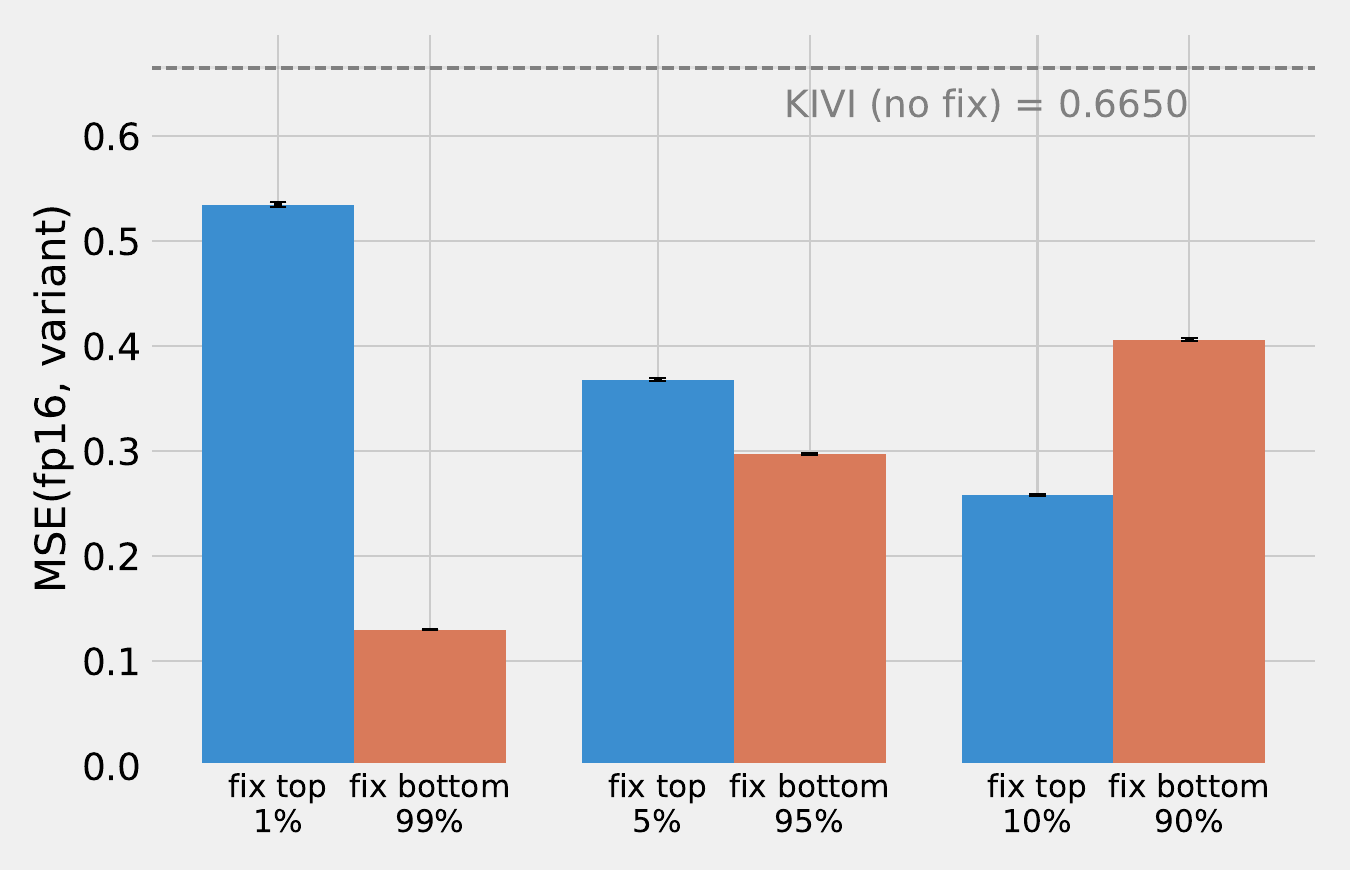}
    \caption{Plot to complement Fig.~\ref{fig:outlier_impact}. It shows much MSE remains, if we replace the largest top k\% errors with the high precision value.  Note that the top 5\% of entries contribute less to MSE than the bottom 95\%, but their end-to-end KL-divergence impact is greater than that of those bottom 95\%.}
    \label{fig:outlier_MSE}
\end{figure}

In Fig.~\ref{fig:outlier_MSE} we show which error quantiles contribute how much to the total MSE of the model. This is best compared with Fig.~\ref{fig:outlier_impact}, where we show the same quantile contributions for end-to-end KL-divergence. Crucially, the 5\% largest errors cause a minority of the MSE, but a majority of the end-to-end KL-divergence. In this sense we can say that fixing outliers is disproportionally important. 

\section{Experimental details}
\label{sec:exp_details}

\paragraph{Models.}
We evaluate four models spanning different capability profiles and parameter scales: Qwen3-4B (\texttt{Qwen/Qwen3-4B}), Llama-3.1-8B-Instruct (\texttt{meta-llama/Llama-3.1-8B-Instruct}), Phi-4 (\texttt{microsoft/phi-4}), and Phi-4-reasoning-plus (\texttt{microsoft/Phi-4-reasoning-plus}). Qwen3-4B is evaluated on all tasks (with reasoning mode activated only for AIME 2024, MATH500 and HumanEval). \hbox{Phi-4} is used for IFEval and Line Retrieval only and Phi-4-reasoning-plus for AIME 2024, MATH-500, and HumanEval.

\paragraph{Quantization.}
All quantized runs use 2-bit KV cache compression with a fixed three-region layout: $S$ sink tokens in FP16 (preserving attention sink behavior), a 2-bit quantized body composed of groups of $G$ tokens each, and $R$ trailing tokens in FP16 corresponding to the most recently generated tokens, which have not yet accumulated enough context to form a full quantization group. We use a classical setting with $G{=}128$, $S{=}128$, $R{=}128$ throughout, except for IFEval where $S{=}32$. Methods that prescribe mixed-precision within groups, such as KVQuant~\citep{hooper2024kvquant}, PolarQuant~\citep{wu2026polarquant}, and TurboQuant~\citep{zandieh2026turboquant}, retain their original per-element precision allocation inside each group. In KVarN, auxiliary parameters (zeropoints and scales) are stored at 8-bit precision.

\paragraph{Decoding.}
We follow the recommended decoding parameters for each model. Qwen3-4B uses thinking mode for AIME 2024, MATH-500 and HumanEval (temp.\ 0.6, top-$p$ 0.95, top-$k$ 20) and non-thinking mode elsewhere (temp.\ 0.7, top-$p$ 0.8, top-$k$ 20). Llama-3.1-8B-Instruct uses temp.\ 0.6, top-$p$ 0.9. Phi-4 decodes greedily. Phi-4-reasoning-plus uses temp.\ 0.8, top-$p$ 0.95, top-$k$ 50 with the official model-card system prompt.

\paragraph{IFEval.}
Full \texttt{google/IFEval} benchmark~\citep{zhou2023instructionfollowingevaluationlargelanguage_ifeval} (541 prompts, \texttt{max\_new\_tokens=1280}).

\paragraph{MATH-500 and AIME 2024.}
MATH-500~\citep{lightman2024lets_math500} (500 problems, \texttt{max\_new\_tokens=8192}); answers extracted from \verb|\boxed{}|. AIME 2024~\citep{aime24} uses the 30 competition problems from \texttt{AI-MO/aimo-validation-aime}, with \texttt{max\_new\_tokens=16384} to accommodate extended chain-of-thought reasoning. Both benchmarks report Avg@3, averaged over 3 independent runs.

\paragraph{HumanEval}
Extended version of HumanEval~\citep{chen2021evaluating_humaneval} (164 problems, \texttt{max\_new\_tokens=16384}), graded by execution with a 30\,s timeout. We report Avg@3.

\paragraph{Line Retrieval.}
Following \cite{li2023how_line-retrieval}, we construct contexts of $L \in \{100, 200, 300, 400, 500, 600\}$ numbered lines, each holding a random 10-character alphanumeric code drawn uniformly from $\{\mathrm{A},\dots,\mathrm{Z}\} \cup \{0,\dots,9\}$. The model is asked to retrieve the code at a randomly chosen line. We use 100 samples per $L$ (\texttt{max\_new\_tokens}$=32$) and report exact-match accuracy. The prompt follows the template:

\begin{quote}\small
\texttt{Below is a list of numbered lines, each with a unique 10-character alphanumeric code.}\\
\texttt{line 1: <$c_1$>}\\
\texttt{line 2: <$c_2$>}\\
\texttt{\ldots}\\[2pt]
\texttt{What is the code on line $k$? Reply with only the 10-character code, nothing else.}
\end{quote}

The predicted answer is the last 10-character alphanumeric token found in the output.

\paragraph{Needle-in-a-Haystack (NIAH).}
We use the benchmark of \cite{kamradt2023needle} with Paul Graham essays as the haystack. The needle is inserted at depth $d \in \{0.1, \ldots, 0.9, 1\}$, where $d$ denotes the fractional position within the haystack, across context lengths spanning 10-100\% of Qwen3-4B's \num{32768}-token window (\num{3112} to \num{31129} tokens, yielding a $10{\times}10$ evaluation grid (\texttt{max\_new\_tokens=128}). The \textit{Static} setting uses the standard quantization protocol, while \textit{Accumulated} uses the pseudo-decode setting introduced in Figure~\ref{fig:pseudo-decode}. The prompt follows the template:

\begin{quote}\small
\textit{You are a helpful assistant. Read the following long passage carefully and remember its contents. I will quiz you about it at the end.}\\[2pt]
\textit{<haystack with needle inserted at depth $d$>}\\[2pt]
\textit{What is the best thing to do in San Francisco?}
\end{quote}

\noindent where the needle is the verbatim sentence: \emph{``The best thing to do in San Francisco is eat a sandwich and sit in Dolores Park on a sunny day''}~\citep{kamradt2023needle}. A response scores 10 if both \emph{sandwich} and \emph{Dolores} appear (case-insensitive), 5 if exactly one, and 0 otherwise.

\paragraph{Baselines.}
KVQuant~\citep{hooper2024kvquant} additionally retains 1\% of $K$ and $V$ channels in FP16. PolarQuant~\citep{wu2026polarquant} requires 4-bit precision for the key cache to yield meaningful results. TurboQuant~\citep{zandieh2026turboquant} is evaluated in a 3-bit K / 3-bit V configuration; as no official implementation is available, we use the community implementation merged into the vLLM framework~\citep{kwon2023efficientvLLM}\footnote{\url{https://github.com/vllm-project/vllm/pull/38479}}, in which the KV cache of the first and last two layers is left unquantized.

\paragraph{Effective memory overhead.}
All bits-per-element figures reported in the tables include auxiliary storage
(scales and zero-points) in addition to the quantized values. For our method, $K$ and $V$ elements are stored at 2~bits with two FP8 scales (one of which absorbs the Sinkhorn normalisation scale into the RTN scale) and
one FP16 zero-point per group of $G{=}128$ elements, giving $2.25$ bits/element.
For KVQuant~\citep{hooper2024kvquant}, the original configuration applies 2-bit
group quantization ($G{=}128$) with FP16 scale and zero-point, plus the top-1\%
of channels retained in FP16, yielding $\approx\!2.4$ bits/element.
For TurboQuant~\citep{zandieh2026turboquant}, the publicly available implementation leaves the KV cache of the first and last two transformer layers unquantized (FP16), making the effective overhead model-dependent; when a single figure is reported it is the average across all layers for that model.

\section{Limitations}
\label{sec:limitations}
Some novel LLM architectures do not require a KV-Cache (e.g. state-space models SSMs \cite{somvanshi2025s4_mamba}). Our method is not suitable for such architectures. 

Some recent models use train-time compression for the KV-Cache in the form of multi-head latent attention (MLA) \cite{liu2024deepseek}. As in other KV-Cache compression methods, it is unclear how such attention mechanisms affect quantization quality. 

End-to-end evaluation on a publicly available serving framework is hindered by the fact that currently such frameworks do not support 2-bit KV-Caches.

\section{Comparison with Eviction Methods}
\label{sec:evict}

In Tab.~\ref{tab:eviction}, we compare against KV cache eviction approaches such as SnapKV~\citep{NEURIPS2024_28ab4182_SnapKV}, PyramidKV~\citep{cai2025pyramidkv}, and KVZip~\citep{kim2026kvzip} on the Line Retrieval task with Llama-3.1-8B-Instruct. Unlike quantization, which compresses the KV cache uniformly across both prefill and generation, these methods apply compression only to the prompt cache and leave the generation cache untouched. This makes them orthogonal to quantization and explains why we restrict the comparison to this appendix. To match the effective memory footprint of our method ($2.3$ bits per element on average), we apply a $7\times$ compression ratio to the prompt cache.

\input{tables/eviction_app}

\section{NiaH: Needle-in-a-Haystack}
\label{sec:niah}
We also evaluate on the Needle-in-a-Haystack task~\citep{kamradt2023needle}. The \textit{Static} setting involves prefill only and thus no error accumulation, making the \textit{Accumulated} pseudo-decode setting (Section~\ref{sec:err_acc}) essential to assess retrieval quality under realistic decoding conditions. Further experimental details are provided in Appendix~\ref{sec:exp_details}. Crucially, the static version of NiaH that is commonly used for KV-Cache compression evaluation is clearly much easier than the accumulated version in pseudo-decode setting.

\begin{figure*}[t]
  \centering
  \includegraphics[width=\textwidth]{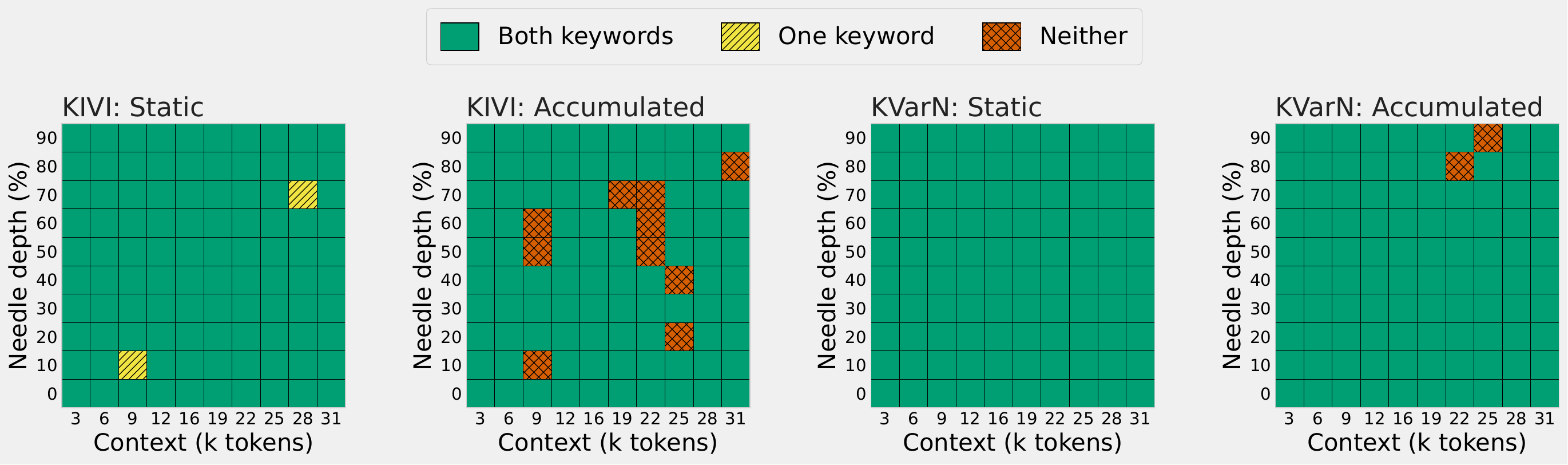}
  \caption{Needle-in-a-Haystack on Qwen3-4B: KIVI vs. \methodname under
  Static and Accumulated prefill. Cells are colored and hatched by
  retrieval outcome (both keywords / one keyword / neither).}
  \label{fig:niah}
\end{figure*}


\section{Variance Normalization Algorithm}
\label{sec:sinq-alg}
We perform the variance normalization on the $K$ and $V$ matrices using Alg.~\ref{alg:sinq}, adapted from \cite{muller2025sinq}.

\begin{algorithm}[t]
\small
\caption{VarN, Variance-Normalization by Alternating Log-Domain Dual-Scale Balancing}
\label{alg:sinq}
\begin{algorithmic}[1]
\Require Batched tiles $\mathbf{T} \in \mathbb{R}^{N \times R \times C}$, Iterations $K$, Limits $c_\text{min}, c_\text{max}$
\Ensure Balanced tiles $\mathbf{T}_{\text{bal}}$, Scales $\mathbf{S}_c \in \mathbb{R}^{N \times 1 \times C}, \mathbf{S}_r \in \mathbb{R}^{N \times R \times 1}$

\Function{Imb}{$\mathbf{X}$}
    \State $\vec{v}_c \gets \text{Var}_{\text{col}}(\mathbf{X}); \quad \vec{v}_r \gets \text{Var}_{\text{row}}(\mathbf{X})$ \Comment{Variances across dims 2 and 1}
    \State \Return $\frac{\max(\vec{v}_c)}{\max(\min(\vec{v}_c), 10^{-8})} + \frac{\max(\vec{v}_r)}{\max(\min(\vec{v}_r), 10^{-8})}$
\EndFunction
\vspace{0.1cm}

\State $\mathbf{L}_c \gets \mathbf{0}_{N \times 1 \times C} ; \quad \mathbf{L}_r \gets \mathbf{0}_{N \times R \times 1}$ \Comment{Initialize log-scales}
\State $\mathbf{C} \gets (\mathbf{T} \oslash \exp(\mathbf{L}_c)) \oslash \exp(\mathbf{L}_r)$ \Comment{Current normalized tiles}
\State $I_{\text{best}} \gets \text{Imb}(\mathbf{C})$
\State $\mathbf{S}_c^* \gets \exp(\mathbf{L}_c) ; \quad \mathbf{S}_r^* \gets \exp(\mathbf{L}_r)$ \Comment{Track best linear scales}

\For{$k \gets 1$ to $K$}
    \State $\mathbf{v}_{\text{col}} \gets \text{clamp}(\text{Var}_\text{row}(\mathbf{C}), c_\text{min}, c_\text{max})$
    \State $\mathbf{L}_c \gets \text{clamp}(\mathbf{L}_c + 0.5 \cdot \log(\mathbf{v}_{\text{col}}), -0.3, 10.0)$
    \State $\mathbf{C} \gets (\mathbf{T} \oslash \exp(\mathbf{L}_c)) \oslash \exp(\mathbf{L}_r)$
    
    \State $\mathbf{v}_{\text{row}} \gets \text{clamp}(\text{Var}_\text{col}(\mathbf{C}), c_\text{min}, c_\text{max})$
    \State $\mathbf{L}_r \gets \text{clamp}(\mathbf{L}_r + 0.5 \cdot \log(\mathbf{v}_{\text{row}}), -0.3, 10.0)$
    \State $\mathbf{C} \gets (\mathbf{T} \oslash \exp(\mathbf{L}_c)) \oslash \exp(\mathbf{L}_r)$
    
    \State $I_{\text{curr}} \gets \text{Imb}(\mathbf{C})$
    \For{$n \gets 1$ to $N$} \Comment{Batched conditional update over $N$ tiles}
        \If{$I_{\text{curr}}[n] \leq I_{\text{best}}[n]$}
            \State $I_{\text{best}}[n] \gets I_{\text{curr}}[n]$
            \State $\mathbf{S}_c^*[n] \gets \exp(\mathbf{L}_c[n]) ; \quad \mathbf{S}_r^*[n] \gets \exp(\mathbf{L}_r[n])$
        \EndIf
    \EndFor
\EndFor

\State \Return $(\mathbf{T} \oslash \mathbf{S}_c^*) \oslash \mathbf{S}_r^*, ~ \mathbf{S}_c^*, ~ \mathbf{S}_r^*$ 
\end{algorithmic}
\end{algorithm}

\section{KVarN Dequantization Overhead}
\label{sec:kvarn_dequant}

We measure the wall-clock cost of dequantizing a full layer of the
quantized KV cache (16 attention heads, head dimension $128$, group size
$128$) for context lengths $4$k-$32$k tokens. The KIVI baseline performs
standard single-scale dequantization. KVarN additionally requires
a per-row second scale $s_2$ from its dual scaling, which we fuse into
the dequant kernel so that no extra HBM round-trip is incurred.

\begin{figure}[t]
  \centering
  \includegraphics[width=0.7\linewidth]{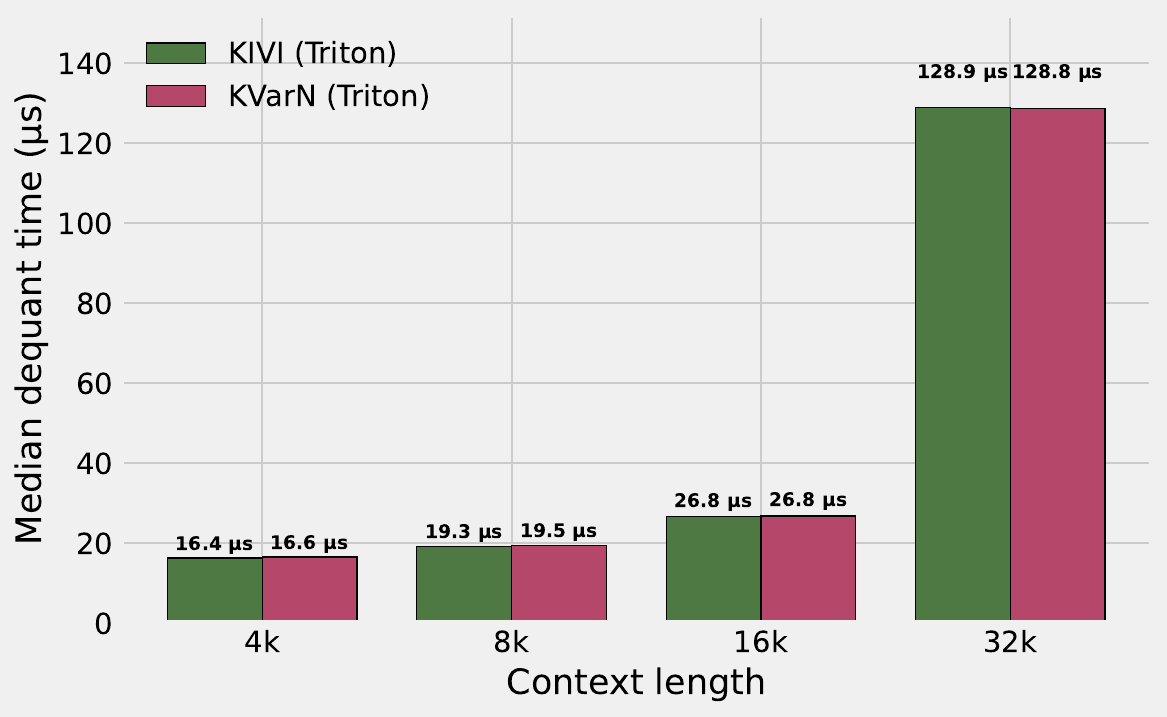}
  \caption{Median dequantization time per call for KIVI (single scale)
  vs.\ KVarN (dual scale, $s_2$ fused into the kernel) across context
  lengths. Lower is better.}
  \label{fig:kvarn_dequant}
\end{figure}

Figure~\ref{fig:kvarn_dequant} reports the median time per call (multiple repeated runs) for both kernels, implemented in Triton on GPU.
Across all context lengths the gap between KVarN and KIVI is at most
$1.4\%$ and within measurement noise at $16$k and $32$k tokens. This
shows that KVarN's dual scaling adds negligible runtime
overhead over the single-scale baseline once $s_2$ is fused into the
dequant kernel.

\section{Computational Cost of Experiments}
\label{sec:computational_cost}
The dominant share of compute cost is due to the experiments on MATH500, AIME24, HumanEval and IFEval. The total cost to reproduce these  are circa 50 GPU days on a GPU with 500 TFLOP at fp16 and 1.8 TB/s memory bandwidth.

\include{checklist}

\end{document}

%% file: tables/reasoning_appendix.tex
\begin{table*}[t]
\centering
\begin{threeparttable}
\small
\setlength{\tabcolsep}{5pt}
\caption{Performance on AIME24 and MATH500. Values are Accuracy / \# Tokens (mean $\pm$ std). Higher accuracy is better; best results per column are highlighted. K/V denotes bits per key/value. UP (Uniform Precision) indicates whether all elements within a given $K$ or $V$ tensor share the same precision ($\checkmark$) or use mixed precision ($\times$).}
\label{tab:math_reasoning_results}
\begin{tabular}{l l c c c c c c c}
\toprule
& & & & & \multicolumn{2}{c}{\textbf{AIME24}} & \multicolumn{2}{c}{\textbf{MATH500}} \\
\cmidrule(lr){6-7} \cmidrule(lr){8-9}
& \textbf{Method} 
& \textbf{K/V} 
& \textbf{bits/elem}
& \textbf{UP}
& \textbf{Acc. (\%)} & \textbf{\# Tokens} 
& \textbf{Acc. (\%)} & \textbf{\# Tokens} \\
\midrule

\multirow{9}{*}{\rotatebox{90}{Qwen3-4B}}
 & \textit{FP16} & 16/16 & 16.0 & $\checkmark$
   & \textit{61.1\tiny$\pm$3.1} & \textit{\num{12477}\tiny$\pm$421} 
   & \textit{82.6\tiny$\pm$0.5} & \textit{\num{3857}\tiny$\pm$23} \\
\cmidrule(lr){2-9}
 & \others{KIVI} & \others{2/2} & \others{2.3} & \others{$\checkmark$} 
   & \others{55.5\tiny$\pm$6.9} & \others{\num{12794}\tiny$\pm$325} 
   & \others{77.8\tiny$\pm$0.5} & \others{\num{3957}\tiny$\pm$7} \\
 & \others{QuaRot} & \others{2/2} & \others{2.3} & \others{$\checkmark$} 
   & \others{56.7\tiny$\pm$3.3} & \others{\num{12732}\tiny$\pm$194} 
   & \others{78.9\tiny$\pm$0.1} & \others{\num{3907}\tiny$\pm$24} \\
 & \others{KVQuant-1\%} & \others{2/2} & \others{2.4} & \others{$\times$}     
   & \others{40.0\tiny$\pm$3.3} & \others{\num{14794}\tiny$\pm$227} 
   & \others{67.5\tiny$\pm$1.5} & \others{\num{5021}\tiny$\pm$45} \\
 & \others{PolarQuant} & \others{4/2} & \others{3.3} & \others{$\checkmark$} 
   & \others{52.2\tiny$\pm$5.8} & \others{\num{13578}\tiny$\pm$361} 
   & \others{71.1\tiny$\pm$1.6} & \others{\num{5259}\tiny$\pm$60} \\
 & \others{TurboQuant\tnote{1}} & \others{3/3} & \others{4.6} & \others{$\checkmark$} 
   & \others{48.9\tiny$\pm$1.9} & \others{\num{13642}\tiny$\pm$261} 
   & \others{77.0\tiny$\pm$0.9} & \others{\num{3917}\tiny$\pm$16} \\
& \others{Kitty} & \others{2/2} & \others{2.4} & \others{$\times$} 
   & \others{53.3\tiny$\pm$8.8} & \others{\num{13123}\tiny$\pm$192} 
   & \others{78.5\tiny$\pm$0.8} & \others{\num{3834}\tiny$\pm$40} \\
 & \textbf{\ours{\methodname (ours)}} 
                & \ours{2/2} & \ours{2.3} & \ours{$\checkmark$}
                & \ours{\textbf{60.0}\tiny$\pm$1.1} & \ours{\num{12408}\tiny$\pm$334} 
                & \ours{\textbf{79.2}\tiny$\pm$0.4} & \ours{\num{3925}\tiny$\pm$38} \\

\midrule

\multirow{9}{*}{\rotatebox{90}{Phi-4-14B}}
 & \textit{FP16} & 16/16 & 16.0 & $\checkmark$
   & \textit{62.2\tiny$\pm$1.6} & \textit{\num{11747}\tiny$\pm$347} 
   & \textit{84.9}\tiny$\pm$0.9 & \textit{\num{2839}\tiny$\pm$35} \\
\cmidrule(lr){2-9}
 & \others{KIVI} & \others{2/2} & \others{2.3} & \others{$\checkmark$} 
   & \others{57.8\tiny$\pm$1.9} & \others{\num{12529}\tiny$\pm$491} 
   & \others{74.4\tiny$\pm$0.8} & \others{\num{3180}\tiny$\pm$27} \\
 & \others{QuaRot} & \others{2/2} & \others{2.3} & \others{$\checkmark$} 
   & \others{58.9\tiny$\pm$1.9} & \others{\num{12284}\tiny$\pm$443} 
   & \others{77.0\tiny$\pm$0.7} & \others{\num{3029}\tiny$\pm$94} \\
 & \others{KVQuant-1\%} & \others{2/2} & \others{2.4} & \others{$\times$}     
   & \others{55.6\tiny$\pm$5.1} & \others{\num{13479}\tiny$\pm$302} 
   & \others{72.3\tiny$\pm$5.1} & \others{\num{3588}\tiny$\pm$36} \\
 & \others{PolarQuant} & \others{4/2} & \others{3.3} & \others{$\checkmark$} 
   & \others{60.0\tiny$\pm$1.9} & \others{\num{12347}\tiny$\pm$52} 
   & \others{75.8\tiny$\pm$1.0} & \others{\num{3023}\tiny$\pm$20} \\
 & \others{TurboQuant\tnote{1}} & \others{3/3} & \others{4.5} & \others{$\checkmark$} 
   & \others{52.2\tiny$\pm$5.0} & \others{\num{13059}\tiny$\pm$223} 
   & \others{74.2\tiny$\pm$0.1} & \others{\num{3250}\tiny$\pm$55} \\
& \others{Kitty} & \others{2/2} & \others{2.4} & \others{$\times$} 
   & \others{55.6\tiny$\pm$11.7} & \others{\num{12226}\tiny$\pm$257} 
   & \others{74.5\tiny$\pm$0.6} & \others{\num{3171}\tiny$\pm$74} \\
 & \textbf{\ours{\methodname (ours)}} 
                & \ours{2/2} & \ours{2.3} & \ours{$\checkmark$}
                & \ours{\textbf{61.7}\tiny$\pm$1.7} & \ours{\num{11598}\tiny$\pm$235}  
                & \ours{\textbf{84.8}\tiny$\pm$0.7} & \ours{\num{2984}\tiny$\pm$67} \\

\bottomrule
\bottomrule
\end{tabular}
\begin{tablenotes}
\item[1]Results obtained using the vLLM implementation.
\end{tablenotes}
\end{threeparttable}
\end{table*}

%% file: tables/humaneval_appendix.tex
\begin{table*}[t]
\centering
\begin{threeparttable}
\small
\setlength{\tabcolsep}{6pt}
\caption{Performance on HumanEval. Values are Accuracy / \# Tokens (mean $\pm$ std over three runs where applicable). Higher accuracy is better; best results per column are highlighted. K/V denotes bits per key/value. UP (Uniform Precision) indicates whether all elements within a given $K$ or $V$ tensor share the same precision ($\checkmark$) or use mixed precision ($\times$).}
\label{tab:humaneval_results}
\begin{tabular}{l l c c c c c c}
\toprule
& & & & \multicolumn{2}{c}{\textbf{Qwen3-4B}} & \multicolumn{2}{c}{\textbf{Phi-4-14B}} \\
\cmidrule(lr){5-6} \cmidrule(lr){7-8}
\textbf{Method} 
& \textbf{K/V} 
& \textbf{bits/elem}
& \textbf{UP}
& \textbf{Accuracy (\%)} & \textbf{\# Tokens}
& \textbf{Accuracy (\%)} & \textbf{\# Tokens} \\
\midrule

\textit{FP16} 
& 16/16 & 16.0 & $\checkmark$
& \textit{88.8\tiny$\pm$1.8} & \textit{\num{2764}\tiny$\pm$280}
& \textit{88.9\tiny$\pm$1.0} & \textit{\num{3015}}\tiny$\pm$52 \\
\cmidrule(lr){1-8}

\others{KIVI}       
& \others{2/2} & \others{2.3} & \others{$\checkmark$}
& \others{86.4\tiny$\pm$1.3} & \others{\num{3292}\tiny$\pm$44}
& \others{74.6\tiny$\pm$2.4} & \others{\num{4929}\tiny$\pm$211} \\

\others{QuaRot}     
& \others{2/2} & \others{2.3} & \others{$\checkmark$}
& \others{86.3\tiny$\pm$1.9} & \others{\num{3255}\tiny$\pm$88}
& \others{87.0\tiny$\pm$2.5} & \others{\num{3216}\tiny$\pm$8} \\

\others{KVQuant-1\%}    
& \others{2/2} & \others{2.4} & \others{$\times$}
& \others{85.2\tiny$\pm$0.7} & \others{\num{4594}\tiny$\pm$60}
& \others{85.6\tiny$\pm$1.3} & \others{\num{4017}\tiny$\pm$122} \\

\others{PolarQuant} 
& \others{4/2} & \others{3.3} & \others{$\checkmark$}
& \others{80.3\tiny$\pm$2.5} & \others{\num{3811}\tiny$\pm$294}
& \others{86.8\tiny$\pm$0.3} & \others{\num{3378}\tiny$\pm$134} \\

\others{TurboQuant\tnote{1}} 
& \others{3/3} & \others{4.6} & \others{$\checkmark$}
& \others{86.2\tiny$\pm$1.9} & \others{\num{3269}\tiny$\pm$144}
& \others{88.0\tiny$\pm$1.4} & \others{\num{3569}\tiny$\pm$76} \\

\others{Kitty} & \others{2/2} & \others{2.4} & \others{$\times$} 
& \others{86.8\tiny$\pm$0.9} & \others{\num{3085}\tiny$\pm$163} 
& \others{82.7\tiny$\pm$0.7} & \others{\num{3697}\tiny$\pm$147} \\

\textbf{\ours{\methodname (ours)}} 
& \ours{2/2} & \ours{2.3} & \ours{$\checkmark$}
& \ours{\textbf{88.4}\tiny$\pm$0.3} & \ours{\num{3192}\tiny$\pm$83}
& \ours{\textbf{88.2}\tiny$\pm$0.4} & \ours{\num{3220}\tiny$\pm$75} \\

\bottomrule
\bottomrule
\end{tabular}
\begin{tablenotes}
\item[1]Results obtained using the vLLM implementation.
\end{tablenotes}
\end{threeparttable}
\end{table*}

%% file: tables/ifeval_main.tex
\begin{table*}[t]
\centering
\begin{threeparttable}
\small
\setlength{\tabcolsep}{5pt}
\caption{Performance on IFEval. Values are Prompt-Level Strict and Loose accuracy (\%). Higher is better; best results per column are highlighted. K/V denotes bits per key/value. UP (Uniform Precision) indicates whether all elements within a given $K$ or $V$ tensor share the same precision ($\checkmark$) or use mixed precision ($\times$).}
\label{tab:ifeval_results_horizontal}
\begin{tabular}{l c c c cc cc cc}
\toprule
& & & & \multicolumn{6}{c}{\textbf{Model}} \\
\cmidrule(lr){5-10}
\textbf{Method} & \textbf{K/V} & \textbf{bits/elem} & \textbf{UP}
& \multicolumn{2}{c}{\textbf{Qwen3-4B}} 
& \multicolumn{2}{c}{\textbf{Llama-3.1-8B}} 
& \multicolumn{2}{c}{\textbf{Phi-4-14B}} \\
\cmidrule(lr){5-6} \cmidrule(lr){7-8} \cmidrule(lr){9-10}
& & & 
& \textbf{Strict} & \textbf{Loose} 
& \textbf{Strict} & \textbf{Loose} 
& \textbf{Strict} & \textbf{Loose} \\
\midrule

\textit{FP16} & 16/16 & 16.00 & $\checkmark$
& \textit{81.0\%} & \textit{84.3\%} 
& \textit{71.1\%} & \textit{76.8\%} 
& \textit{63.6\%} & \textit{69.3\%} \\

\midrule

\others{KIVI} & \others{2/2} & \others{2.3} & \others{$\checkmark$}
& \others{80.3\%} & \others{83.2\%} 
& \others{70.9\%} & \others{74.5\%} 
& \others{60.6\%} & \others{67.7\%} \\

\others{Hadamard} & \others{2/2} & \others{2.3} & \others{$\checkmark$}
& \others{79.3\%} & \others{82.8\%} 
& \others{70.8\%} & \others{75.0\%} 
& \others{62.6\%} & \others{68.2\%} \\

\others{KVQuant-1\%} & \others{2/2} & \others{2.4} & \others{$\times$}
& \others{76.9\%} & \others{80.8\%} 
& \others{57.7\%} & \others{61.4\%} 
& \others{55.1\%} & \others{61.2\%} \\

\others{PolarQuant} & \others{4/2} & \others{3.3} & \others{$\checkmark$}
& \others{79.1\%} & \others{82.6\%} 
& \others{69.5\%} & \others{75.0\%} 
& \others{62.5\%} & \others{68.4\%} \\

\others{TurboQuant\tnote{1}} & \others{3/3} & \others{4.6} & \others{$\checkmark$}
& \others{79.2\%} & \others{81.7\%} 
& \others{66.5\%} & \others{71.0\%} 
& \others{57.7\%} & \others{62.7\%} \\

\others{Kitty} & \others{2/2} & \others{2.4} & \others{$\times$}
& \others{78.0\%} & \others{81.9\%} 
& \others{70.2\%} & \others{75.2\%} 
& \others{63.2\%} & \others{69.0\%} \\

\textbf{\ours{\methodname (ours)}} & \ours{2/2} & \ours{2.3} & \ours{$\checkmark$}
& \ours{\textbf{80.4\%}} & \ours{\textbf{83.4\%}} 
& \ours{\textbf{71.0\%}} & \ours{\textbf{76.5\%}} 
& \ours{\textbf{63.4\%}} & \ours{\textbf{69.2\%}} \\

\bottomrule
\bottomrule
\end{tabular}
\begin{tablenotes}
\item[1]Results obtained using the vLLM implementation.
\end{tablenotes}
\end{threeparttable}
\end{table*}

%% file: tables/line_retrieval_main.tex
\begin{table*}[t]
\centering
\begin{threeparttable}
\small
\setlength{\tabcolsep}{3pt}
\caption{Performance across context lengths (number of lines). Values are accuracy (\%). Higher is better; best results per column are highlighted. K/V denotes bits per key and value. UP (Uniform Precision) indicates whether all elements within a given $K$ or $V$ tensor share the same precision ($\checkmark$) or use mixed precision ($\times$).}
\label{tab:kv_quant_results}
\begin{tabular}{l l c c c c c c c c c}
\toprule
& & & & & \multicolumn{6}{c}{\textbf{Number of lines}} \\
\cmidrule(lr){6-11}
\textbf{Model} & \textbf{Method} 
& \textbf{K/V}
& \textbf{bits/elem}
& \textbf{UP}
& \textbf{100} & \textbf{200} & \textbf{300} & \textbf{400} & \textbf{500} & \textbf{600} \\
\midrule

\multirow{8}{*}{Qwen3-4B}
 & \textit{FP16} & 16/16 & 16.0 & $\checkmark$
   & \textit{100\%} & \textit{99\%} & \textit{98\%} & \textit{98\%} & \textit{96\%} & \textit{90\%} \\
\cmidrule(lr){2-11}
 & \others{KIVI} & \others{2/2} & \others{2.3} & \others{$\checkmark$} & \others{98\%} & \others{97\%} & \others{88\%} & \others{84\%} & \others{84\%} & \others{74\%} \\
 & \others{Hadamard} & \others{2/2} & \others{2.3} & \others{$\checkmark$} & \others{98\%} & \others{98\%} & \others{91\%} & \others{\textbf{98\%}} & \others{92\%} & \others{83\%} \\
 & \others{KVQuant-1\%} & \others{2/2} & \others{2.4} & \others{$\times$} & \others{93\%} & \others{80\%} & \others{82\%} & \others{69\%} & \others{57\%} & \others{57\%} \\
 & \others{PolarQuant} & \others{4/2} & \others{3.3} & \others{$\checkmark$} & \others{93\%} & \others{86\%} & \others{83\%} & \others{91\%} & \others{82\%} & \others{84\%} \\
 & \others{TurboQuant\tnote{1}} & \others{3/3} & \others{4.6} & \others{$\checkmark$} & \others{99\%} & \others{99\%} & \others{\textbf{98\%}} & \others{95\%} & \others{94\%} & \others{\textbf{85\%}} \\
 & \others{Kitty} & \others{2/2} & \others{2.4} & \others{$\times$} & \others{\textbf{100\%}} & \others{99\%} & \others{93\%} & \others{94\%} & \others{95\%} & \others{79\%} \\
 & \textbf{\ours{\methodname (ours)}} 
   & \ours{2/2} & \ours{2.3} & \ours{$\checkmark$}
   & \ours{\textbf{100\%}} & \ours{\textbf{99\%}} & \ours{\textbf{98\%}} & \ours{97\%} & \ours{\textbf{97\%}} & \ours{\textbf{85\%}} \\

\midrule

\multirow{8}{*}{Llama-3.1-8B}
 & \textit{FP16} & 16/16 & 16.0 & $\checkmark$
   & \textit{100\%} & \textit{99\%} & \textit{99\%} & \textit{96\%} & \textit{93\%} & \textit{92\%} \\
\cmidrule(lr){2-11}
 & \others{KIVI} & \others{2/2} & \others{2.3} & \others{$\checkmark$} & \others{98\%} & \others{97\%} & \others{90\%} & \others{84\%} & \others{80\%} & \others{83\%} \\
 & \others{Hadamard} & \others{2/2} & \others{2.3} & \others{$\checkmark$} & \others{\textbf{99\%}} & \others{97\%} & \others{94\%} & \others{90\%} & \others{84\%} & \others{85\%} \\
 & \others{KVQuant-1\%} & \others{2/2} & \others{2.4} & \others{$\times$} & \others{92\%} & \others{88\%} & \others{77\%} & \others{68\%} & \others{58\%} & \others{54\%} \\
 & \others{PolarQuant} & \others{4/2} & \others{3.3} & \others{$\checkmark$} & \others{59\%} & \others{72\%} & \others{76\%} & \others{77\%} & \others{67\%} & \others{58\%} \\
 & \others{TurboQuant\tnote{1}} & \others{3/3} & \others{4.8} & \others{$\checkmark$} & \others{98\%} & \others{95\%} & \others{96\%} & \others{88\%} & \others{79\%} & \others{83\%} \\
 & \others{Kitty} & \others{2/2} & \others{2.4} & \others{$\times$} & \others{98\%} & \others{\textbf{98\%}} & \others{95\%} & \others{87\%} & \others{80\%} & \others{87\%} \\
 & \textbf{\ours{\methodname (ours)}} 
   & \ours{2/2} & \ours{2.3} & \ours{$\checkmark$}
   & \ours{\textbf{99\%}} & \ours{\textbf{98\%}} & \ours{\textbf{98\%}} & \ours{\textbf{95\%}} & \ours{\textbf{91\%}} & \ours{\textbf{89\%}} \\

\midrule

\multirow{8}{*}{Phi-4-14B}
 & \textit{FP16} & 16/16 & 16.0 & $\checkmark$
   & \textit{100\%} & \textit{100\%} & \textit{100\%} & \textit{98\%} & \textit{100\%} & \textit{95\%} \\
\cmidrule(lr){2-11}
 & \others{KIVI} & \others{2/2} & \others{2.3} & \others{$\checkmark$} & \others{95\%} & \others{96\%} & \others{88\%} & \others{88\%} & \others{91\%} & \others{82\%} \\
 & \others{Hadamard} & \others{2/2} & \others{2.3} & \others{$\checkmark$} & \others{99\%} & \others{\textbf{99\%}} & \others{97\%} & \others{\textbf{97\%}} & \others{91\%} & \others{94\%} \\
 & \others{KVQuant-1\%} & \others{2/2} & \others{2.4} & \others{$\times$} & \others{94\%} & \others{80\%} & \others{86\%} & \others{82\%} & \others{74\%} & \others{67\%} \\
 & \others{PolarQuant} & \others{4/2} & \others{3.3} & \others{$\checkmark$} & \others{98\%} & \others{\textbf{99\%}} & \others{69\%} & \others{\textbf{97\%}} & \others{80\%} & \others{85\%} \\
 & \others{TurboQuant\tnote{1}} & \others{3/3} & \others{4.5} & \others{$\checkmark$} & \others{61\%} & \others{51\%} & \others{55\%} & \others{53\%} & \others{57\%} & \others{56\%} \\
 & \others{Kitty} & \others{2/2} & \others{2.4} & \others{$\times$} & \others{99\%} & \others{98\%} & \others{97\%} & \others{93\%} & \others{96\%} & \others{83\%} \\
 & \textbf{\ours{\methodname (ours)}} 
   & \ours{2/2} & \ours{2.3} & \ours{$\checkmark$}
   & \textbf{\ours{100\%}} & \textbf{\ours{99\%}} & \textbf{\ours{99\%}} & \textbf{\ours{97\%}} & \textbf{\ours{98\%}} & \textbf{\ours{95\%}} \\
\bottomrule
\bottomrule
\end{tabular}
\begin{tablenotes}
\item[1]Results obtained using the vLLM implementation.
\end{tablenotes}
\end{threeparttable}
\end{table*}

%% file: tables/eviction_app.tex
\begin{table*}[t]
\centering
\begin{threeparttable}
\small
\setlength{\tabcolsep}{3pt}
\caption{Performance across context lengths (number of lines). Values are accuracy (\%). Higher is better; best results per column are highlighted. K/V denotes bits per key/value. The equivalent bits/elem reflects the effective compression ratio. UP (Uniform Precision) indicates whether all elements within a given $K$ or $V$ tensor share the same precision ($\checkmark$) or use mixed precision ($\times$).}
\label{tab:eviction}
\begin{tabular}{l l c c c c c c c c c}
\toprule
& & & & & \multicolumn{6}{c}{\textbf{Number of lines}} \\
\cmidrule(lr){6-11}
\textbf{Model} & \textbf{Method} 
& \textbf{K/V}
& \makecell{\textbf{equivalent}\\\textbf{bits/elem}}
& \textbf{UP}
& \textbf{100} & \textbf{200} & \textbf{300} & \textbf{400} & \textbf{500} & \textbf{600} \\
\midrule

\multirow{6}{*}{Llama-3.1-8B}
 & \textit{FP16} & 16/16 & 16.0 & $\checkmark$
   & \textit{100\%} & \textit{99\%} & \textit{99\%} & \textit{94\%} & \textit{93\%} & \textit{91\%} \\
\cmidrule(lr){2-11}
 & \others{SnapKV ($7\times$)} & \others{16/16} & \others{2.3} & \others{$\checkmark$}
   & \others{77\%} & \others{87\%} & \others{90\%} & \others{90\%} & \others{88\%} & \others{85\%} \\
 & \others{PyramidKV ($7\times$)} & \others{16/16} & \others{2.3} & \others{$\checkmark$}
   & \others{61\%} & \others{88\%} & \others{87\%} & \others{88\%} & \others{88\%} & \others{87\%} \\
 & \others{KVZip ($7\times$)} & \others{16/16} & \others{2.3} & \others{$\checkmark$}
   & \others{80\%} & \others{92\%} & \others{86\%} & \others{88\%} & \others{89\%} & \others{86\%} \\
 & \textbf{\ours{\methodname (ours)}} 
   & \ours{2/2} & \ours{2.3} & \ours{$\checkmark$}
   & \ours{\textbf{100\%}} & \ours{\textbf{99\%}} & \ours{\textbf{96\%}} & \ours{\textbf{91\%}} & \ours{\textbf{90\%}} & \ours{\textbf{89\%}} \\

\bottomrule
\end{tabular}
\end{threeparttable}
\end{table*}